\documentclass[runningheads,a4paper]{llncs}

\usepackage{ifxetex}
\ifxetex%
  \usepackage{fontspec}
  \usepackage{polyglossia}
  \setmainlanguage{english}
\else
  \usepackage[utf8]{inputenc}
  \usepackage[T1]{fontenc}
  \usepackage[english]{babel}
\fi

\usepackage{amssymb}
\usepackage{graphicx}
\usepackage{url}
\usepackage{float}
\usepackage{amsmath}
\usepackage{graphicx}
\usepackage{wrapfig}
\usepackage{booktabs}
\usepackage{multirow}
\usepackage{lipsum}
\usepackage{csquotes}
\usepackage{fancyhdr}
\usepackage{lastpage}
\usepackage[all]{nowidow}
\usepackage[inline]{enumitem}
\usepackage[usenames,dvipsnames]{xcolor}

\makeatletter
\def\Hline{%
\noalign{\ifnum0=`}\fi\hrule \@height 1pt \futurelet
\reserved@a\@xhline}
\makeatother

\title{Hibikino-Musashi@Home\\2024 Team Description Paper}

\author{
Kosei Isomoto	
\and Akinobu Mizutani
\and Fumiya Matsuzaki
\and Hikaru Sato
\and Ikuya Matsumoto
\and Kosei Yamao
\and Takuya Kawabata
\and Tomoya Shiba
\and Yuga Yano
\and Atsuki Yokota
\and Daiju Kanaoka
\and Hiromasa Yamaguchi
\and Kazuya Murai
\and Kim Minje
\and Lu Shen
\and Mayo Suzuka
\and Moeno Anraku
\and Naoki Yamaguchi
\and Satsuki Fujimatsu
\and Shoshi Tokuno
\and Tadataka Mizo
\and Tomoaki Fujino
\and Yuuki Nakadera
\and Yuka Shishido
\and Yusuke Nakaoka
\and Yuichiro Tanaka
\and Takashi Morie
\and Hakaru Tamukoh
}
\authorrunning{Kosei Isomoto et al.}
\institute{
Kyushu Institute of Technology\\
The University of Kitakyushu\\
\email{hma@brain.kyutech.ac.jp} \\
\url{https://www.brain.kyutech.ac.jp/~hma/}
}

\begin{document}
\maketitle

%%%%%%%%%%%%%%%%%%%%%%%%%%%%%%%%%%%%%%%%%%%%%%%%%%%%%%%%%%%%%%%%%%%%%%%%%%%%%%%%%%%%
%
% Abstract
%
%%%%%%%%%%%%%%%%%%%%%%%%%%%%%%%%%%%%%%%%%%%%%%%%%%%%%%%%%%%%%%%%%%%%%%%%%%%%%%%%%%%%

\begin{abstract}
  This paper provides an overview of the techniques employed by Hibikino-Musashi@Home, which intends to participate in the domestic standard platform league. The team has developed a dataset generator for training a robot vision system and an open-source development environment running on a Human Support Robot simulator. 
  The large language model powered task planner selects appropriate primitive skills to perform the task requested by users. The team aims to design a home service robot that can assist humans in their homes and continuously attends competitions to evaluate and improve the developed system.
\end{abstract}

%%%%%%%%%%%%%%%%%%%%%%%%%%%%%%%%%%%%%%%%%%%%%%%%%%%%%%%%%%%%%%%%%%%%%%%%%%%%%%%%%%%%

\section{Introduction}

Hibikino-Musashi@Home (HMA)  is a robot development team comprising students at the Kyushu Institute of Technology and the University of Kitakyushu in Japan. The team was founded in 2010 and has participated in the RoboCup@Home JapanOpen in the open platform league (OPL), domestic standard platform league (DSPL), and Simulation-DSPL.  It has recurrently participated in the RoboCup@Home league since 2017 and will participate in RoboCup 2024 to present the outcomes of their latest developments and research. In addition to the RoboCup, the team participated in the World Robot Challenge (WRC) 2018 and 2020 as well as in the service robotics category of the partner robot challenge (real space). HMA focuses on the development of robot vision systems, particularly dataset generation systems for the training of object-recognition systems. It also develops libraries for primitive tasks, including object recognition, grasping point estimation, and navigation. Task planning is their latest topic of interest, which uses a large language model (LLM) to plan a task by selecting primitive tasks in a dynamic environment.

%%%%%%%%%%%%%%%%%%%%%%%%%%%%%%%%%%%%%%%%%%%%%%%%%%%%%%%%%%%%%%%%%%%%%%%%%%%%%%%%%%%%

\section{System Overview}
\begin{figure}[bt]
\begin{center}
\includegraphics[scale=0.35]{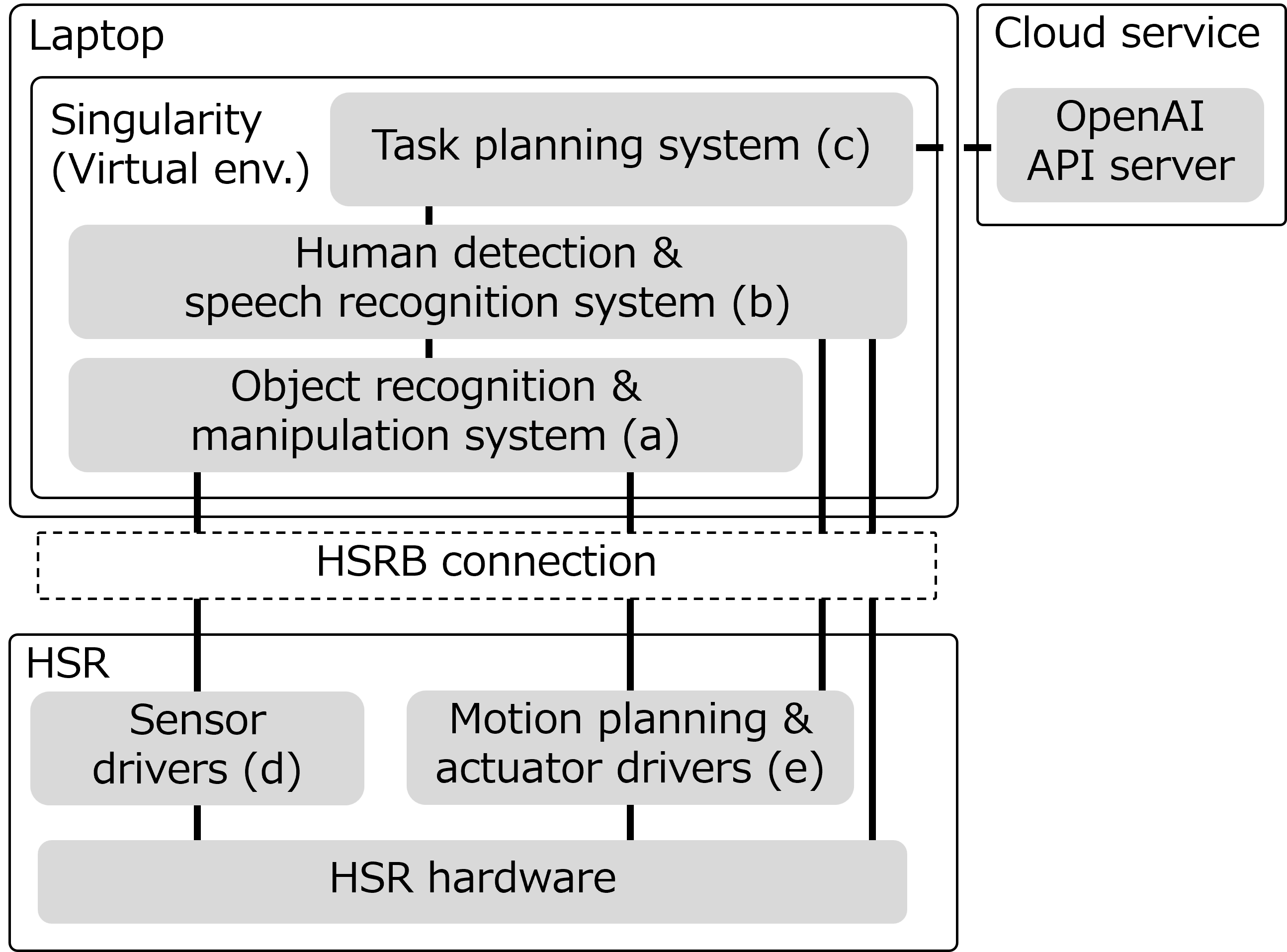} % TODO:図の変更
\caption{Block diagram providing an overview of the  HSR system. [HSR, Human Support Robot; ROS, robot operating system]}
\label{fig:softOverview}
\end{center}
\end{figure}

We use an external laptop (ThinkPad X1 Extreme Gen5) mounted on the back of the TOYOTA Human Support Robot (HSR) to compute systems require high computational power. The computer inside the HSR is also used to run basic HSR libraries, such as sensor drivers, motion planners, and actuator drivers. Most of the system is complete within the mounted laptop and internal HSR computer except the LLM API server.

Figure \ref{fig:softOverview} presents an overview of the HMA's software systems for HSR \cite{yamamoto2019robomech}. 
Our system is on Singularity \cite{singularity}, a virtual environment tool built on a laptop mounted on the HSR. The system comprises an object recognition and manipulation system (a), a human and speech recognition system (b), and a task planning system (c), which are all connected to the sensor and actuator drivers (d) and (e) on the HSR's internal computer, respectively.
The task planning system (c) communicates with the OpenAI API server \cite{OpenAI_API} via the Internet.

%%%%%%%%%%%%%%%%%%%%%%%%%%%%%%%%%%%%%%%%%%%%%%%%%%%%%%%%%%%%%%%%%%%%%%%%%%%%%%%%%%%%
%
% Perception (Environment Understanding)
% 1. Object Recognition
% 2. Grasping Point Estimation
% 3. Speech Recognition
% 4. Human Detection
% 5. Semantic Map
%
%%%%%%%%%%%%%%%%%%%%%%%%%%%%%%%%%%%%%%%%%%%%%%%%%%%%%%%%%%%%%%%%%%%%%%%%%%%%%%%%%%%%
\section{Perception}
\subsection{Object Recognition}
The object-recognition system is a crucial component of robot systems. We have adopted a strategy that leverages both the YOLOv8 \cite{Jocher_YOLO_by_Ultralytics_2023} and Language Segment-Anything \cite{Lang_SAM}. We select the best system for each task depending on the target objects.

\subsubsection{YOLO}
\begin{figure}[tb]
\begin{center}
  \includegraphics[width=0.5\columnwidth]{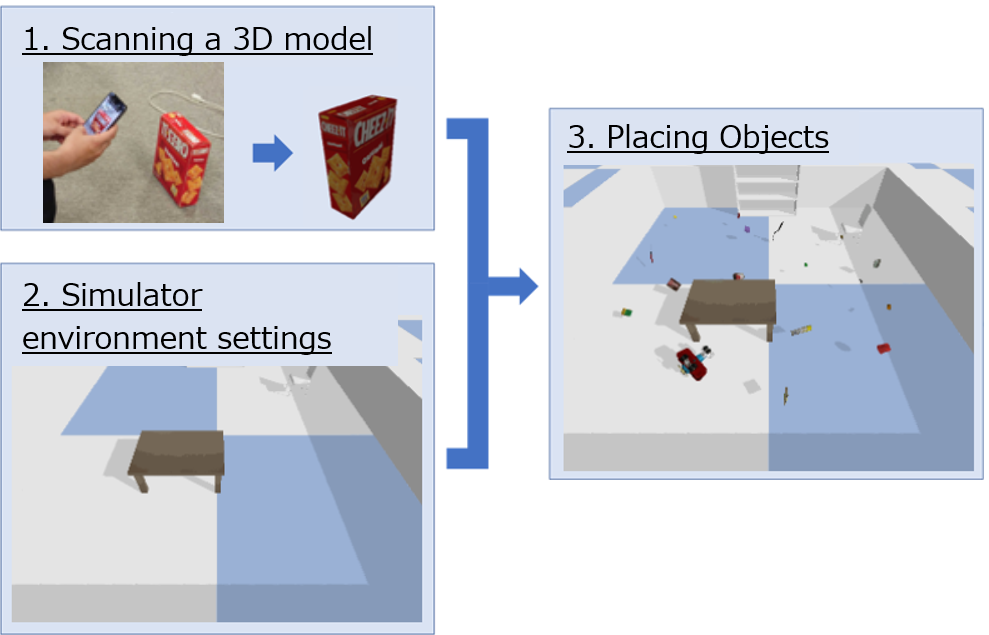} % TODO:図の変更
  \caption{Placing scanned objects on a 3D simulator.}
  \label{fig:annotation}
\end{center}
\end{figure}

YOLOv8 has low latency with accuracy that is sufficient to recognize the known objects in the competition. 
We fine-tune the YOLO system to recognize the objects used in the competition using enormous training dataset generated by the 3D simulation system based on the PyBullet \cite{coumans2016pybullet} simulator \cite{ono2022ar}. 
As shown in Fig. \ref{fig:annotation}, to generate the training dataset, first, we create a 3D model of each object by using a smartphone with LiDAR sensors. The scanned 3D objects are spawned in the 3D environment, and the objects as well as the environment are shot from various angles to create a dataset with 500,000 images in under 2 h using a six-core CPU simultaneously. The light conditions, placement of the furniture, and texture of the background (floor, wall, and ceiling) are changed to randomize a domain at each shot. The annotation data for the training data can be generated automatically by the system; no annotation process by humans is required.

\subsubsection{Language Segment-Anything}
\begin{figure}[tb]
\begin{center}
    \includegraphics[width=0.8\columnwidth]{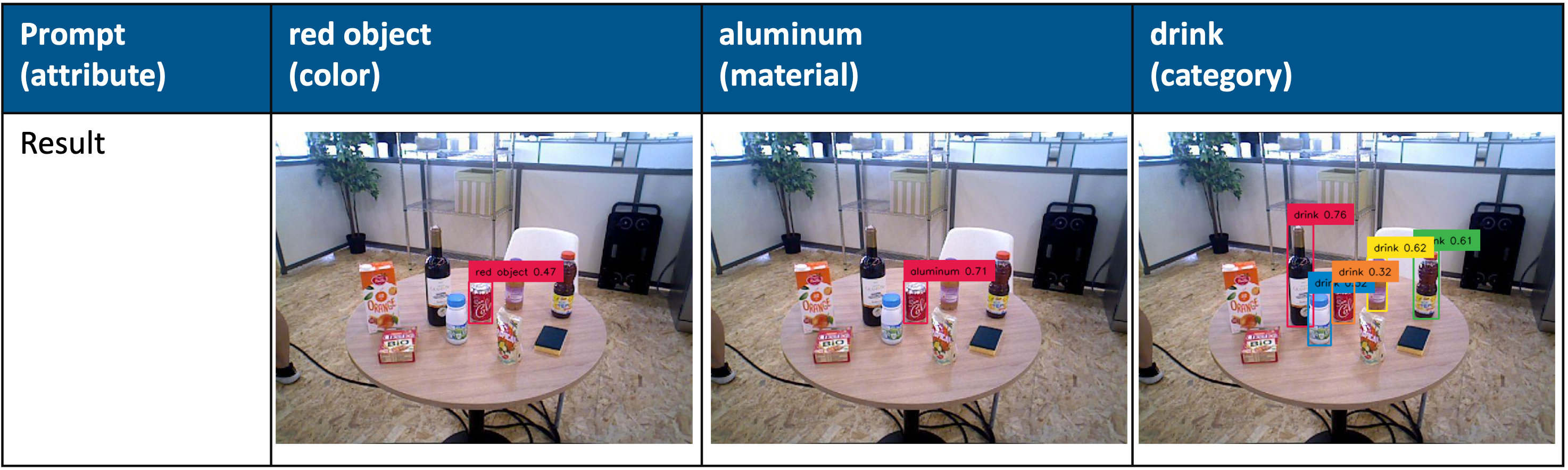}
    \caption{Example of the prompts for Language Segment-Anything.}
    \label{fig:object_features}
  \end{center}
  \end{figure}

Language Segment-Anything \cite{Lang_SAM} is a combination of Grounding DINO \cite{G_DINO} and Segment Anything \cite{SAM}. Grounding DINO is a recognition system that can be tuned by providing text prompts. Segment anything model (SAM) can also output the segmentation mask of the object without any fine-tuning.
In this method, the human has to provide the text prompt and the recognition accuracy depends on this prompt. However, the system can be tuned quickly without fine-tuning; thus, the known objects available on the competition site can be recognized without extensive preparation. The target object can be detected by giving different types of prompts such as color, material and category as shown in Fig. \ref{fig:object_features}. Humans are required to choose and tune the prompt by checking the recognition result in advance.

% \begin{figure}[tb]
% \begin{center}
%   \includegraphics[width=0.5\columnwidth]{./images/langsam.png}
%   \caption{Overview of the LangSAM.}
%   \label{fig:langsam}
% \end{center}
% \end{figure}

%%%%%%%%%%%%%%%%%%%%%%%%%%%%%%%%%%%%%%%%%%%%%%%%%%%%%%%%%%%%%%%%%%%%%%%%%%%%%%%%%%%%

% \subsection{Grasping Point Estimation}

% As shown in Fig. \ref{fig:grasp_point}, the robot system calculates the grasping point by using 3D point clouds. The procedure of the proposed method is as follows: 
% \begin{itemize}
%   \setlength{\leftskip}{1.0cm}
%   \item [Step (a)] Obtain a 3D point cloud.
%   \item [Step (b)] Remove planes using segmentation.
%   \item [Step (c)] Extract information regarding each object using clustering.
%   \item [Step (d)] Calculate the size and direction of each object and estimate the grasping direction.
% \end{itemize}

% By using this method, the robot can swiftly estimate the grasping point using the visual information of the object without any pre-acquired knowledge.

% \begin{figure}[tb]
%   \begin{center}
%     \includegraphics[width=0.5\columnwidth]{./images/grasp_point.jpg}
%     \caption{Overview of the grasping point estimation.}
%     \label{fig:grasp_point}
%   \end{center}
% \end{figure}
%%%%%%%%%%%%%%%%%%%%%%%%%%%%%%%%%%%%%%%%%%%%%%%%%%%%%%%%%%%%%%%%%%%%%%%%%%%%%%%%%%%%

\subsection{Speech Recognition}

We have built a real-time speech recognition system.
HSR acquires the speaker's voice in frames and transmits it sequentially to an external laptop computer.
Our system uses Voice Activity Detection (VAD) \cite{pyvad2016} to determine when to start and stop recording.
Post recording, our system automatically starts speech recognition using only the speech section's audio data.
We use Whisper \cite{radford2022robust}, a high-accuracy speech recognition method.

% \begin{figure}[tb]
%   \begin{center}
%     \includegraphics[width=0.75\columnwidth]{./images/speech_recognition/whisper_realtime_speech_recognition.jpg}
%   \caption{Real-time speech recognition system for more natural HRI.}
%   \label{fig:speech_recog}
%   \end{center}
% \end{figure}

%%%%%%%%%%%%%%%%%%%%%%%%%%%%%%%%%%%%%%%%%%%%%%%%%%%%%%%%%%%%%%%%%%%%%%%%%%%%%%%%%%%%

\subsection{Human Detection}
\begin{figure}[tb]
  \begin{center}
    \includegraphics[width=0.75\columnwidth]{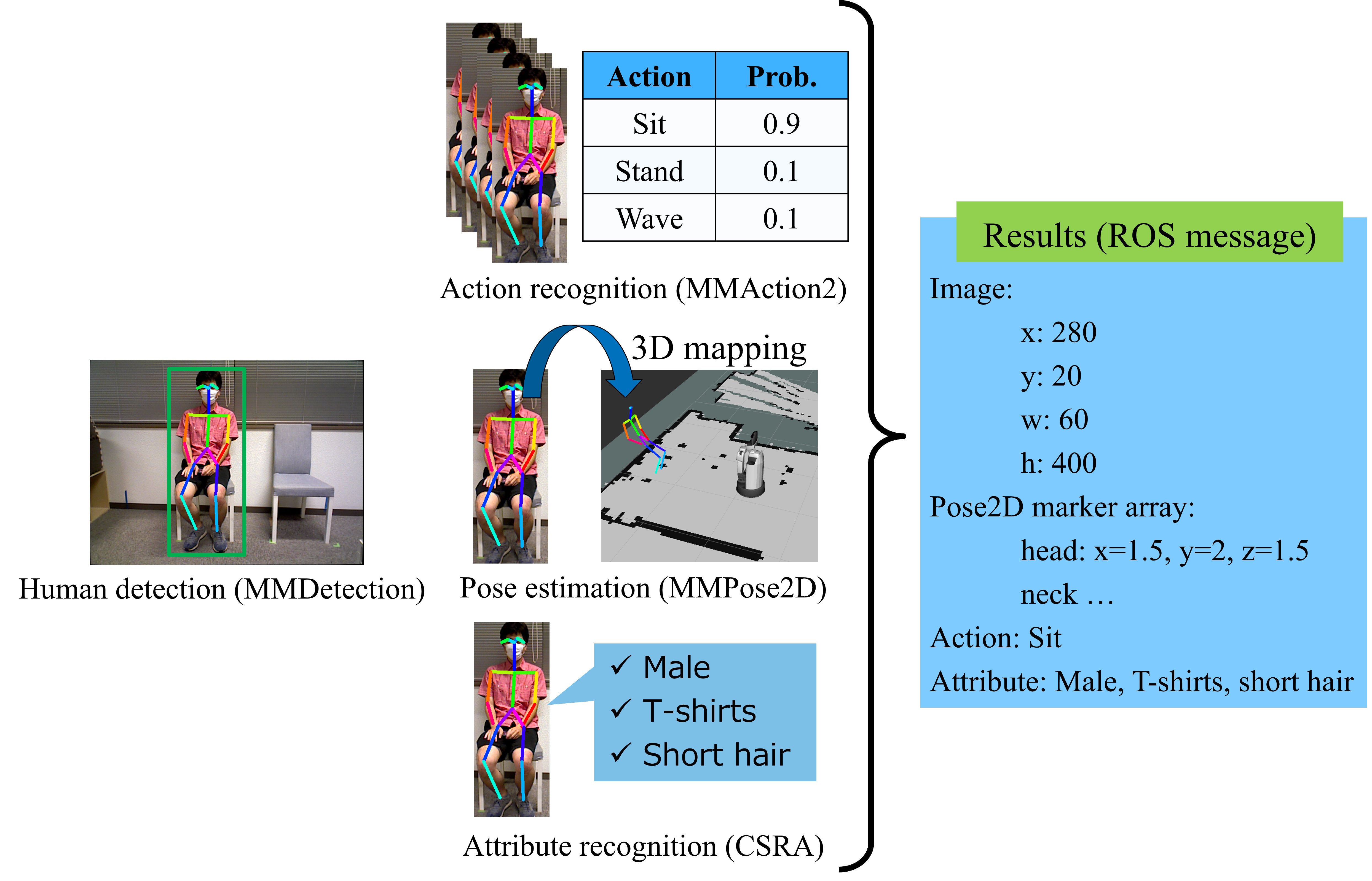} % TODO:素材変更
  \caption{Placing scanned objects on a 3D simulator.}
  \label{fig:mm_overview}
  \end{center}
\end{figure}

We use the MM Libraries provided by OpenMMLab \cite{openmmlab} for human recognition. 
Figure \ref{fig:mm_overview} demonstrates our human recognition system.
We use MMDetection to detect humans and obtain cropped images.
We detect human key points through MMPose and recognize human action using MMAction2.
In addition, we use Class-Specific Residual Attention (CSRA) \cite{Zhu2021CSRA} to obtain human attributes, such as gender, hairstyle, and clothing.

\subsection{Human Tracking}
We use YOLOv7 \cite{wang2023yolov7} and StrongSORT \cite{Du2023StrongSORT} for human tracking tasks such as \textit{Carry My Luggage} and \textit{General Purpose Service Robot (GPSR)}. In these tasks, the robot must to follow the human to the destination in a crowded environment.

Humans can be detected by YOLOv7; however, it cannot identify a human uniquely. To track a particular human, we use StrongSORT, a multi-object tracking(MOT) system.
% When using MOT, there are problems inherent to MOT, such as missing associations and missing detections.
% However, tasks such as \textit{Carry My Luggage} require continuous recognition of specific persons.
% Therefore, we employed StrongSORT and modified the system to return only the bounding box of the first person recognized as information.

%%%%%%%%%%%%%%%%%%%%%%%%%%%%%%%%%%%%%%%%%%%%%%%%%%%%%%%%%%%%%%%%%%%%%%%%%%%%%%%%%%%%
\subsection{Semantic Map}
\begin{figure}[tb]
  \begin{center}
    \includegraphics[width= 0.5\columnwidth]{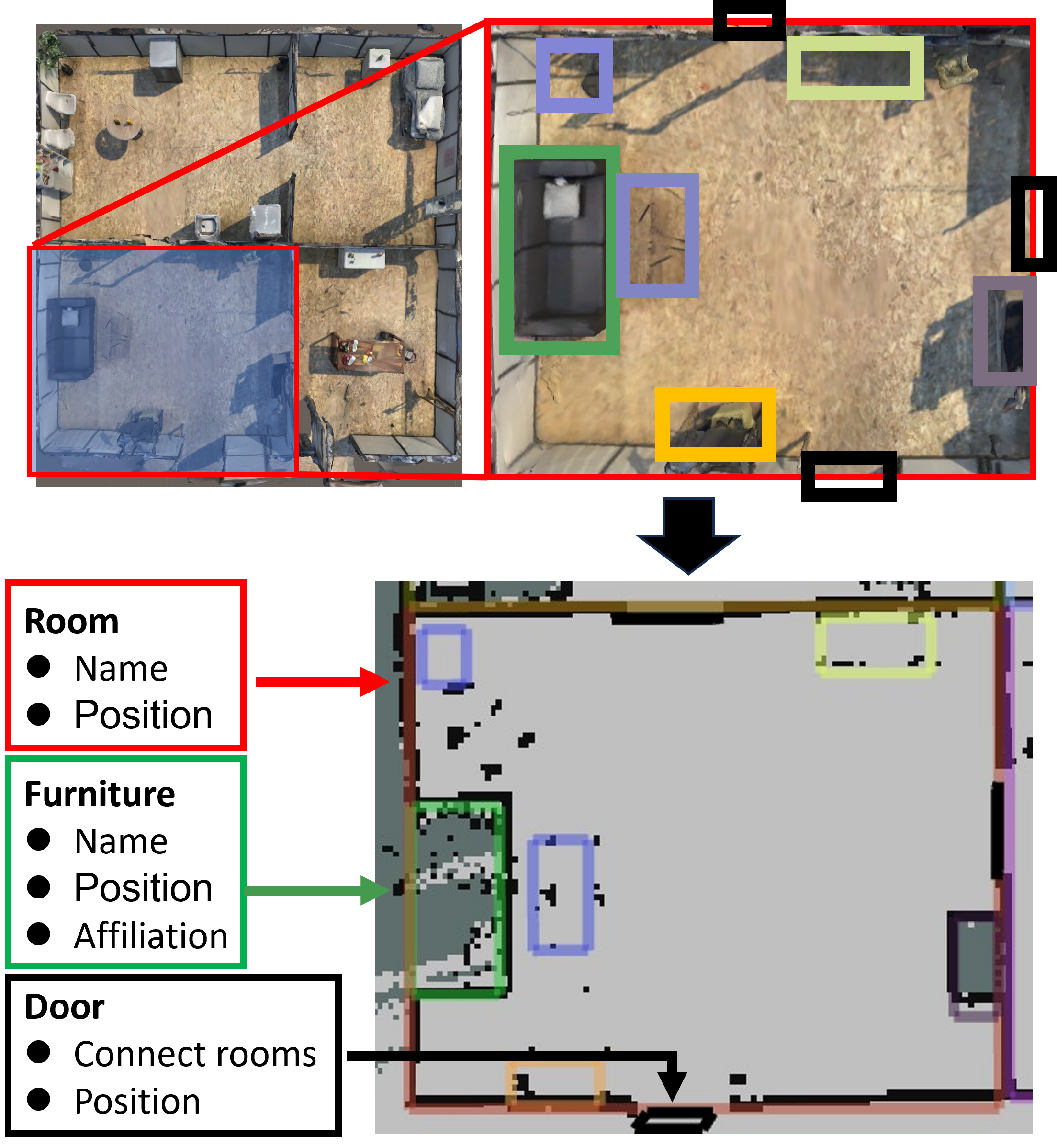}
    \caption{Example of a semantic map}
    \label{fig:semantic-map}
  \end{center}
\end{figure}

A semantic map is required for the autonomous task execution of the service robots. 
As shown in Fig. \ref{fig:semantic-map}, we add the semantic information about the room, furniture, and door to a pre-acquired environment map created using Real-Time Appearance-Based Mapping (RTAB-map) \cite{mathieu2019rtabmap}.
Each room and furniture has a label with the name and location information using an array of two-dimensional (2D) coordinates representing the vertex information of the contours.
Moreover, the furniture has information regarding the room in which it is located, and each door has information regarding the room to which it is connected. 
Before the task execution, we define areas of the room and the positions of the door and furniture to input to a configuration file by humans.
A robot utilizes that information to not only determine the semantic location of the robot or humans in an arena but also plan the path to the target position.

The path planning algorithm using semantic maps works as follows. First, the robot receives the 2D coordinates of the destination location. Then, it determines the destination room and the room of the robot's current position using the inside/outside determination method based on the outer product. Finally, it calculates the shortest path from the robot's current position to the destination. This calculation includes the Euclidean distance from the robot's position to the door, between doors that the robot passes through to reach the desired room, and from the door to the specified furniture. The algorithm can change the path by setting the door as passable. This allows the robot to replan its path by setting the door as impassable after detecting that it is closed.

% In \textit{GPSR} and \textit{Enhanced GPSR (EGPSR)}, the robot plans the path to the destination based on the semantic and the target place information, which is included in a command instead of the 2D coordinates of the destination location.

%%%%%%%%%%%%%%%%%%%%%%%%%%%%%%%%%%%%%%%%%%%%%%%%%%%%%%%%%%%%%%%%%%%%%%%%%%%%%%%%%%%%
\section{Task Planning}
\begin{figure}[tb]
  \begin{center}
  \includegraphics[width=0.9\columnwidth]{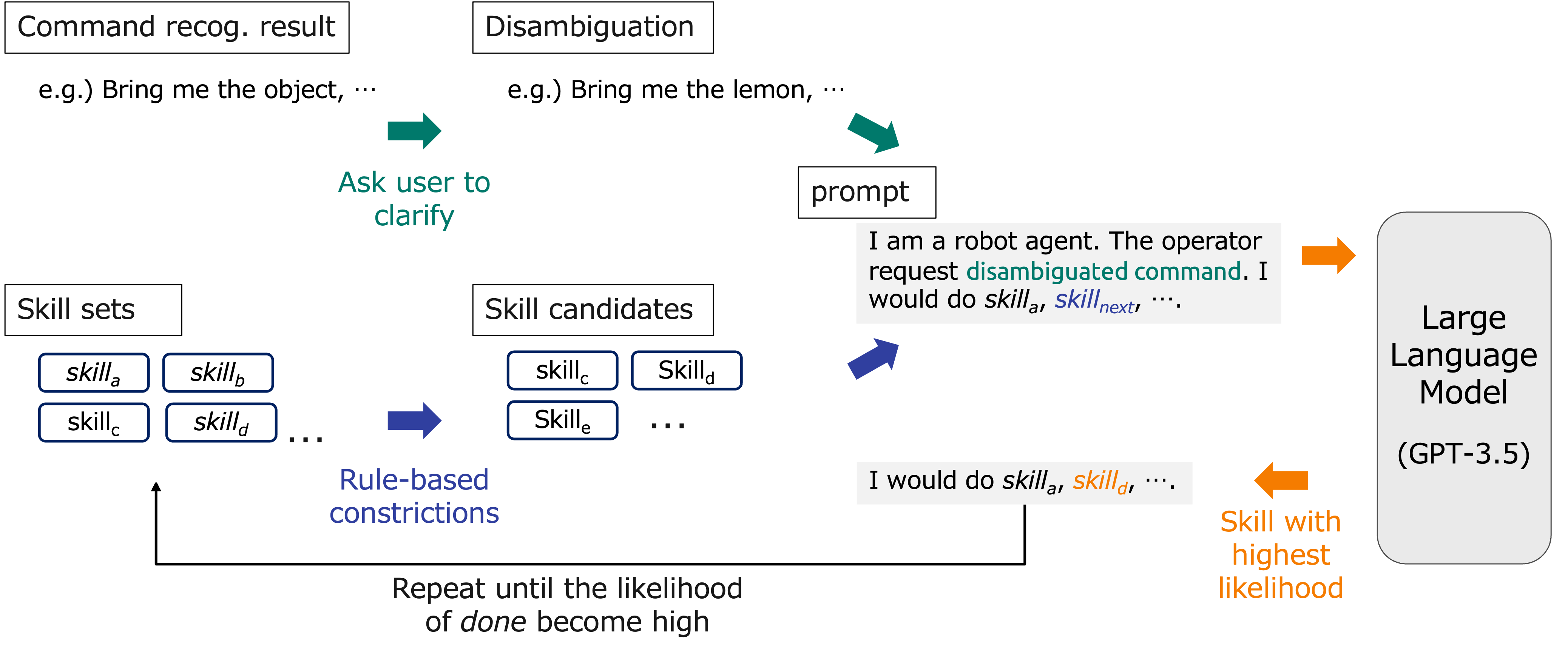} % TODO:図の変更
  \caption{Overview of the task planning}
  \label{fig:task_planning}
  \end{center}
\end{figure}

A task planning system is required to accomplish various requests from a user in real-world environments. As shown in Fig. \ref{fig:task_planning}, the proposed system plans the task using command recognition results and skill sets based on the SayCan system \cite{ahn2022i}. 

The voice requests from the users are converted into text. The given command may have an ambiguous word, for example, \textit{Bring me the object}. In this case, the system asks the user for the proper noun of the ambiguous word/phrase. Then, based on the user's answer, the system replaces the ambiguous word with the proper noun. 
The skill set is the set of available skills for the robot. The possible skills are selected based on the rule-based constrictions, for example, the same skill is not repeated consecutively, and \texttt{grasp} is after \texttt{find\_obj}. The likelihood of each possible skill is calculated by the LLM, and the skill with the highest likelihood is selected as the next robot's task until the likelihood of \textit{done} becomes the highest.

To plan the task, the system requires common and environment-specific knowledge. For example, to bring a bottle of water to the family, the possible location can be thought of using common knowledge and the actual location depends on each home environment. To acquire the environment-specific information, we have proposed a brain-inspired memory acquisition model with hippocampus functions \cite{mizutani2021,mizutani2023}. This model is designed to be implemented on low-power consumption hardware such as FPGAs and dedicated chips \cite{kawashima2022}. We think this model is necessary for the robot to work in a home environment and the latest research activity will be demonstrated in the \textit{Open Challenge}.

% \begin{figure}[tb]
%   \begin{center}
%   \includegraphics[width=0.75\columnwidth]{./images/command_rec03.png} % TODO:図の変更
%   \caption{The overview of the command recognition}
%   \label{fig:command_rec}
%   \end{center}
% \end{figure}

% \begin{table}[tb]
%   \centering
%   \caption {Skill sets used in the proposed systems}
%   \label{tab:skill_set}
%   \begin{tabular}{c|l}
%     \hline
%     Key  & Value \\ \hline \hline
%     \texttt{move}  &  go to the \{PLACE\}\\
%     \texttt{follow}  &  follow the target  \\
%     \texttt{find\_obj}  &  find the \{OBJECT\} \{PREPOSITION\} the \{PLACE\}  \\
%     \texttt{find\_person}  &  find \{PERSON\}  \\
%     \texttt{observe\_obj}  &  look at the \{PLACE\} to check objects  \\
%     \texttt{observe\_person}  &  look at the \{PLACE\} to check people  \\
%     \texttt{grasp\_obj}  &  grasp the \{OBJECT\}  \\
%     \texttt{put}  &  take the \{OBJECT\} to the \{PLACE\}  \\
%     \texttt{pass\_obj}  &  pass the \{OBJECT\}  \\
%     \texttt{answer\_question}  &  answer a question  \\
%     \texttt{say}  &  say \{\} \\
%     \texttt{done}  &  done  \\ \hline
%   \end{tabular}
% \end{table}

%%%%%%%%%%%%%%%%%%%%%%%%%%%%%%%%%%%%%%%%%%%%%%%%%%%%%%%%%%%%%%%%%%%%%%%%%%%%%%%%%%%%

\section{Reusability}
We have published our development workspace used in RoboCup 2021 on the open-source HSR simulators \cite{hsrb_robocup_dspl_docker} on GitHub\footnote{\texttt{hma\_wrs\_sim\_ws} (\url{https://github.com/Hibikino-Musashi-Home/hma_wrs_sim_ws})}. It includes documentation and sample programs with motion-synthesis and object-recognition libraries. This simulator workspace enables us to develop the robot system even without the physical HSR.  It can also be used for the testing and evaluation of the robot system. Currently, we are working on the development of an open-source development workspace for physical HSR with the virtual environment on Singularity.

%%%%%%%%%%%%%%%%%%%%%%%%%%%%%%%%%%%%%%%%%%%%%%%%%%%%%%%%%%%%%%%%%%%%%%%%%%%%%%%%%%%%

% \vspace{-0.3cm}
\section{Conclusions}
This paper describes the techniques for creating an intelligence system for home-service robots.  The automatic dataset generation system is essential to train the visual system of a service robot in a limited time. The task planning system is sufficiently powerful to create the robot's action by the human's spoken request, and necessary primitive tasks are continuously developed by the team to enhance the functions of the home service robot.

%%%%%%%%%%%%%%%%%%%%%%%%%%%%%%%%%%%%%%%%%%%%%%%%%%%%%%%%%%%%%%%%%%%%%%%%%%%%%%%%%%%%
% \vspace{-0.3cm}
\section*{Acknowledgment}
This paper is based on results obtained from a project, JPNP16007, commissioned by the New Energy and Industrial Technology Development Organization (NEDO).
This paper is also supported by Joint Graduate School Intelligent Car, Robotics \& AI, Kyushu Institute of Technology student project, YASKAWA electric corporation project, JSPS KAKENHI grant number 22K17968, and JST SPRING grant number JPMJSP2154.
% \newpage
%\vspace{-0.3cm}
\bibliography{ref}
\bibliographystyle{splncs04}

\newpage
\section*{Appendix 1: Robot's Software Description}
\begin{wrapfigure}[20]{r}{0.25\textwidth}
        \centering
        \includegraphics[width=0.25\textwidth]{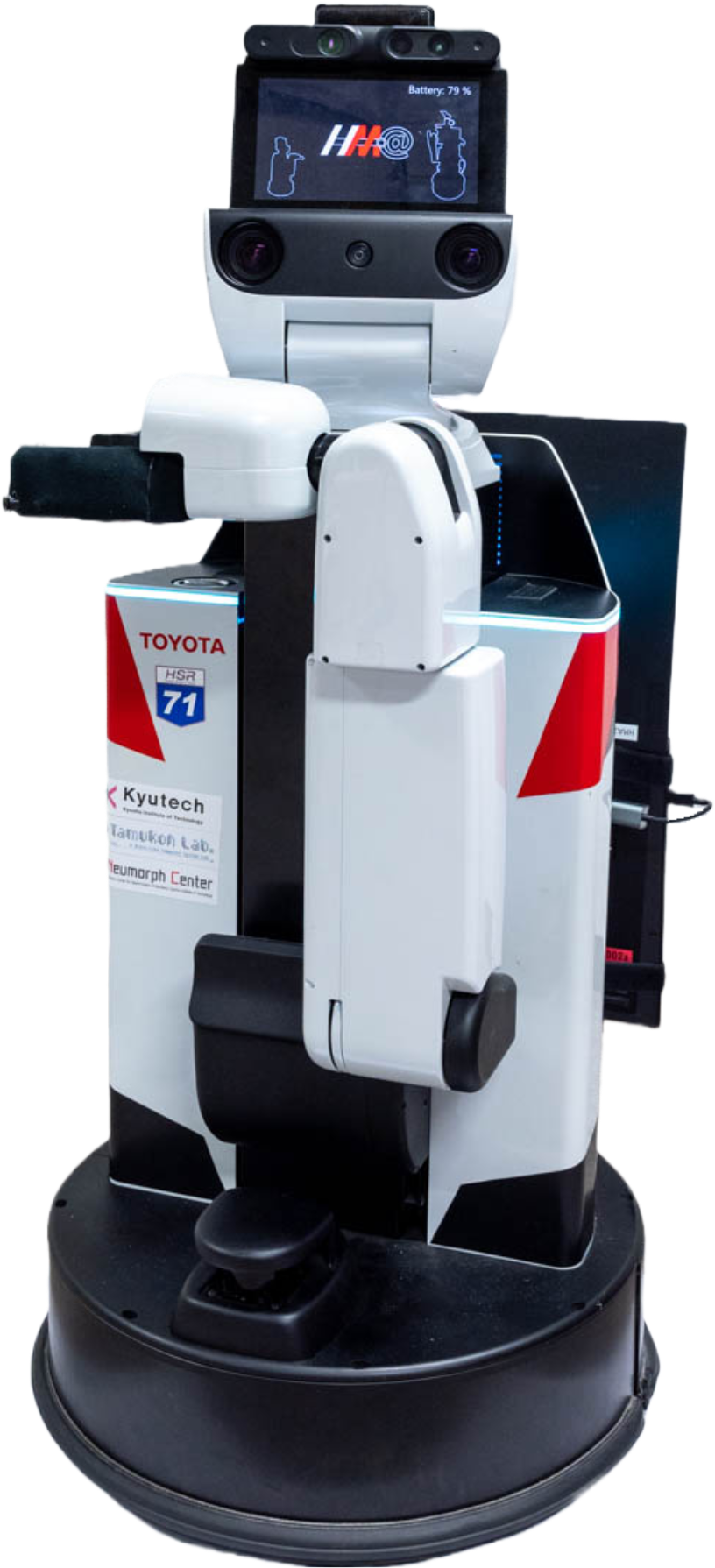}
        \caption{HSR}%
        \label{fig:hsr}
\end{wrapfigure}

The following is the software stack of our robot system, as shown in Fig. \ref{fig:hsr}.
\begin{itemize}
	\item OS: Ubuntu 20.04
	\item Middleware: ROS Noetic
	\item State management: SMACH (ROS)
	\item Speech recognition: Whisper \cite{radford2022robust}
	\item Object detection: YOLO and Language Segment-Anything \cite{Jocher_YOLO_by_Ultralytics_2023,Lang_SAM}
	\item Human detection/action recognition:
		\begin{itemize}
			\item MM Libraries (MMDetection, MMPose, MMAction2) \cite{openmmlab}
			\item YOLO and StrongSORT \cite{wang2023yolov7,Du2023StrongSORT}
		\end{itemize}
	\item Attribute recognition: CSRA \cite{Zhu2021CSRA}
	% \item Face detection: Convolutional Neural Network.
	\item SLAM: rtabmap \cite{mathieu2019rtabmap} (ROS)
	\item Path planning: move\_base (ROS)
\end{itemize}

The following are the specifications of the laptop mounted on our HSR.
\begin{itemize}
        \item Model name: ThinkPad X1 Extreme Gen5
        \item CPU: Intel Core i9-12900H
        \item RAM: 32GB
        \item GPU: NVIDIA GeForce RTX 3080Ti (16GB)
\end{itemize}

\section*{Appendix 2: Competition results}

\begin{table}[t]
\begin{center}
\caption{Results of the recent competitions. [DSPL, domestic standard-platform league; JSAI, Japanese Society for Artificial Intelligence; METI, Ministry of Economy, Trade and Industry (Japan); OPL, open-platform league; RSJ, Robotics Society of Japan]}
\label{tab:result}
\begin{tabular}{l|l} \hline
	\multicolumn{1}{c|}{Competition} & \multicolumn{1}{c}{Result} \\ \hline \hline
%	Japan & RoboCup Japan Open 2017 Aichi & @Home DSPL 2nd \\
%        							&& @Home OPL 3rd \\ \hline

	RoboCup 2017 Nagoya & {\bf @Home DSPL 1st} \\
                                & @Home OPL 5th \\ \hline

        RoboCup JapanOpen 2018 Ogaki & @Home DSPL 2nd \\
                                & \textbf{@Home OPL 1st} \\
                                & JSAI Award \\ \hline

        RoboCup 2018 Montreal & \textbf{@Home DSPL 1st} \\
                                & P\&G Dishwasher Challenge Award \\ \hline

        World Robot Challenge 2018 & {\bf Service Robotics Category} \\
                                & \textbf{Partner Robot Challenge Real Space 1st} \\
                                & METI Minister's Award, RSJ Special Award \\ \hline

        RoboCup 2019 Sydney & @Home DSPL 3rd \\ \hline

        RoboCup JapanOpen 2019 Nagaoka &\textbf{@Home DSPL 1st} \\
                                        & \textbf{@Home OPL 1st} \\ \hline
        RoboCup JapanOpen 2020 & @Home Simulation Technical Challenge 2nd \\
                                        & \textbf{@Home DSPL 1st} \\
                                        & @Home DSPL Technical Challenge 2nd \\
                                        & \textbf{@Home OPL 1st} \\
                                        & \textbf{@Home OPL Technical Challenge 1st} \\
                                        & @Home Simulation DSPL 2nd \\ \hline
        RoboCup Worldwide 2021 & @Home DSPL 2nd \\
                                        & \textbf{@Home Best Open Challenge Award 1st} \\
                                        & \textbf{@Home Best Test Performance: } \\
                                        & \textbf{    Go, Get It! 1st} \\
                                        & \textbf{@Home Best Go, Get It! 1st} \\ \hline
        World Robot Challenge 2020 & \bf{Service Robotics Category} \\
                                        & \bf{Partner Robot Challenge Real Space 1st} \\ \hline
        RoboCup Asia-Pacific 2021 Aichi Japan & \textbf{@Home DSPL 1st}\\
                                        & \textbf{@Home OPL 1st} \\ \hline
        RoboCup JapanOpen 2021 & \textbf{@Home DSPL 1st} \\
                                        & \textbf{@Home DSPL Technical Challenge 1st} \\
                                        & @Home OPL 2nd \\
                                        & \textbf{@Home OPL Technical Challenge 1st} \\ \hline
        RoboCup 2022 Bangkok & @Home DSPL 3rd\\ 
                                        & @Home Best Open Challenge Award\\
                                        & @Home Robo-host \\
                                        & (Party-Host highest score in Stage I tasks)\\ \hline
        RoboCup JapanOpen 2022 Tokyo & \textbf{@Home DSPL 1st} \\
                                        & @Home DSPL Technical Challenge 2nd \\
                                        & @Home OPL 2nd \\
                                        & \textbf{@Home OPL Technical Challenge 1st} \\ \hline
        RoboCup JapanOpen 2023 Shiga & \textbf@Home DSPL 3rd \\
                                        & \textbf{@Home DSPL Open Challenge 1st} \\
                                        & @Home OPL 2nd \\
                                        & \textbf{@Home OPL Open Challenge 1st} \\ \hline
        RoboCup 2023 Bordeaux & @Home DSPL 2nd\\ \hline
\end{tabular}
\end{center}
\end{table}

Table \ref{tab:result} shows the results achieved by our team in the recent competitions.
We have been participating in the RoboCup and World Robot Challenge for several years. Our team has won several prizes and academic awards. \par

\section*{Appendix 3: Links}

\begin{itemize}
  % ToDo: Update YouTube Link
  \item Team Video \\ \url{https://youtu.be/VKKz-PcQsvc} \\
  \item Team Website \\ \url{https://www.brain.kyutech.ac.jp/~hma} \\
  \item GitHub \\ \url{https://github.com/Hibikino-Musashi-Home} \\
  \item Facebook \\ \url{https://www.facebook.com/HibikinoMusashiAthome} \\
  \item YouTube \\ \url{https://www.youtube.com/@hma_wakamatsu}
\end{itemize}

\end{document}